\documentclass[10pt,twocolumn,letterpaper]{article}

\usepackage[pagenumbers]{cvpr} 

\usepackage[dvipsnames]{xcolor}

\definecolor{cvprblue}{rgb}{0.21,0.49,0.74}
\usepackage[breaklinks,colorlinks,citecolor=cvprblue]{hyperref}
\usepackage{hyperref}
\usepackage{url}
\usepackage{graphicx}
\usepackage{float}
\usepackage{algorithm}  
\usepackage{algorithmicx}
\usepackage{algpseudocode}
\usepackage{natbib}
\usepackage{amsmath}
\usepackage{hyperref}
\usepackage{booktabs}
\usepackage{caption}
\usepackage{subcaption}
\usepackage{color}

\usepackage{indentfirst}

\algnotext{EndFor}
\algnotext{EndIf}
\algnotext{EndProcedure}

\def\paperID{15039} 
\def\confName{CVPR}
\def\confYear{2024}

\vspace{-15pt}
\title{OrthCaps: An Orthogonal CapsNet with Sparse Attention Routing and Pruning}

\vspace{-15pt}
\author{
    Xinyu Geng\textsuperscript{1}  \hspace{15pt}
    Jiaming Wang\textsuperscript{1}  \hspace{15pt}
    Jiawei Gong\textsuperscript{1}  \hspace{15pt}
    Yuerong Xue\textsuperscript{1}  \\
    Jun Xu\textsuperscript{1\dag}  \hspace{15pt}
    Fanglin Chen\textsuperscript{1\dag}  \hspace{15pt}
    Xiaolin Huang\textsuperscript{2}  \\
    \textsuperscript{1} Harbin Institute of Technology, Shenzhen\\
    \textsuperscript{2} Shanghai Jiao Tong University\\
    {\tt\small \{22s153095, 21s153144\}@stu.hit.edu.cn,
    \{jiawei\_gong, xyuerong\}@163.com, }\\
    {\tt\small
    xujunqgy@hit.edu.cn,
    chenfanglin@gmail.com,
    xiaolinhuang@sjtu.edu.cn
    }
    }
\begin{document}
\hyphenpenalty=4000

\maketitle
\begin{abstract}
    Redundancy is a persistent challenge in Capsule Networks (CapsNet),
    leading to high computational costs and parameter counts. 
    Although previous works have introduced pruning after the initial capsule layer,
    dynamic routing's fully connected nature and non-orthogonal weight matrices reintroduce redundancy in deeper layers.
    Besides, dynamic routing requires iterating to converge, further increasing computational demands.
    In this paper, 
    we propose an Orthogonal Capsule Network (OrthCaps) to reduce redundancy, improve routing performance and decrease parameter counts.
    Firstly, an efficient pruned capsule layer is introduced to discard redundant capsules. 
    Secondly, dynamic routing is replaced with orthogonal sparse attention routing, eliminating the need for iterations and fully connected structures.  
    Lastly, weight matrices during routing are orthogonalized to sustain low capsule similarity,
    which is the first approach to introduce orthogonality into CapsNet as far as we know.
    Our experiments on baseline datasets affirm the efficiency and robustness of OrthCaps in classification tasks, 
    in which ablation studies validate the criticality of each component. 
    Remarkably, OrthCaps-Shallow outperforms other Capsule Network benchmarks on four datasets, 
    utilizing only 110k parameters – a mere 1.25\% of a standard Capsule Network's total.
    To the best of our knowledge, it achieves the smallest parameter count among existing Capsule Networks.
    Similarly, OrthCaps-Deep demonstrates competitive performance across four datasets, 
    utilizing only 1.2\% of the parameters required by its counterparts. 
    The code is available at \url{https://github.com/ornamentt/Orthogonal-Capsnet}.

    \renewcommand{\thefootnote}{}
    \footnotetext{\textsuperscript{\dag} Corresponding authors.}
    \renewcommand{\thefootnote}{\arabic{footnote}}
\end{abstract}    
\vspace*{-10pt}

\section{Introduction}
\label{sec:intro}

\begin{figure}[t]
   \vspace{-5pt}
   \centering
   \includegraphics[scale=0.5]{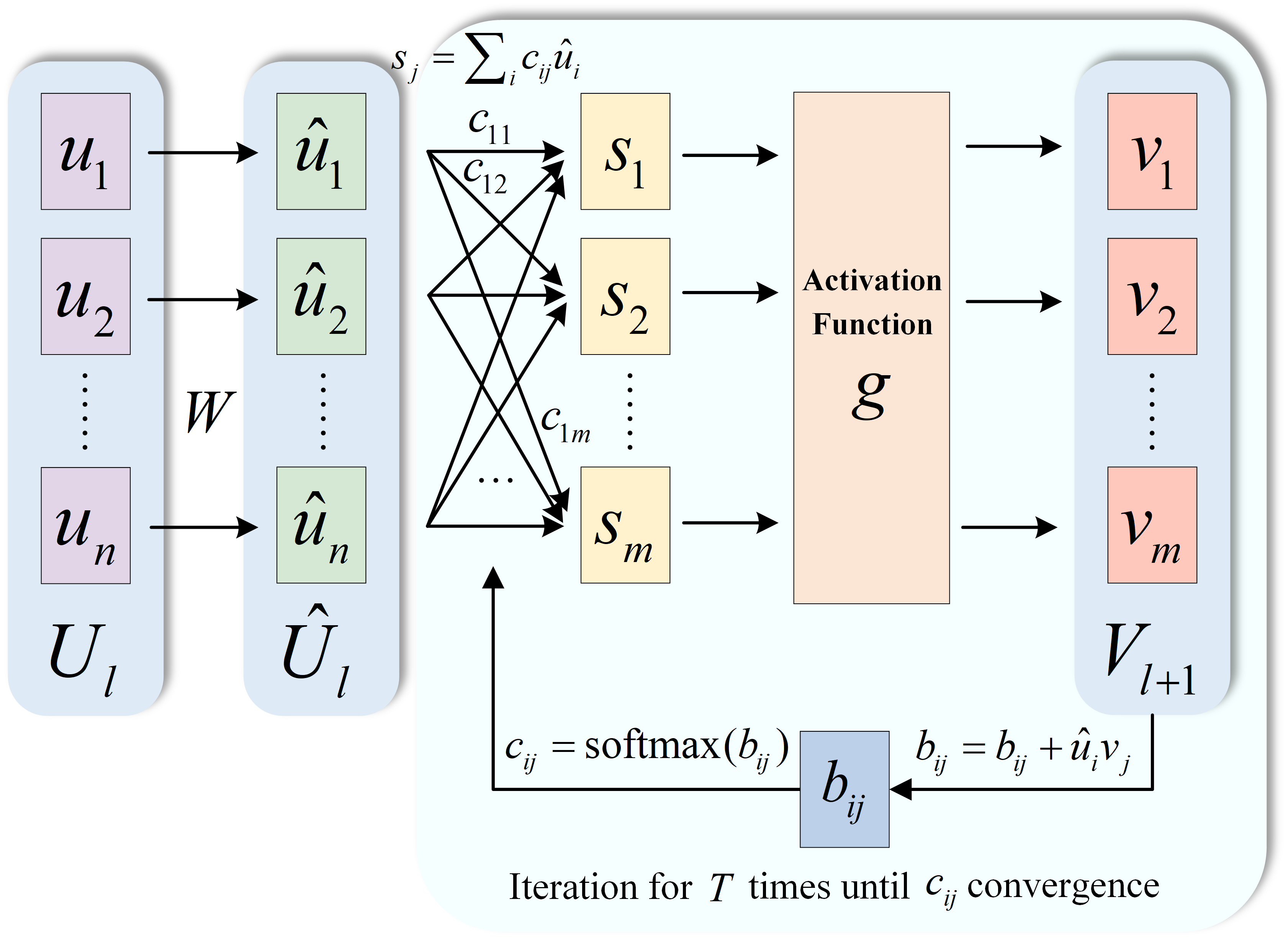}
   \caption{Dynamic routing mechanism. 
   $u_i,v_j$ are the lower-level capsule, and higher-level capsule, respectively.
   $W$ is the weight matrix to produce the pose prediction $\hat{u}_i$ of $u_i$ for next level.
   $b_{ij}$ is a temporary variable to calculate the coupling coefficient $c_{ij}$.
   }
   \label{dynamic}
   \vspace*{-17pt}
\end{figure}

\indent 
While convolutional Neural Networks (CNNs) excel in computer vision tasks, certain challenges remain,
which include information loss in pooling layers, low robustness, and poor spatial feature correlation \cite{sabour2017dynamic,hinton2018matrix}. 
To address these limitations, 
Capsule Network (CapsNet) was proposed, using capsule vectors instead of traditional neurons.
In CapsNet, each capsule vector's length represents the presence probability of specific entities in the input image, 
and its direction encodes the captured features \cite{sabour2017dynamic}.
This setup allows the capsule vectors to capture features related to corresponding entities. 
CapsNet's architecture includes a primary capsule extraction layer, a digit capsule layer, dynamic routing, and class-conditioned reconstruction. 
As a key component of CapsNet, dynamic routing aligns lower-level capsules with higher-level ones,
which is described in \cref{dynamic}. 
First, lower-level capsules $U_l$ (in matrix form) predict poses $\hat U_l$ for higher-level capsules $V_{l+1}$ via weight matrix $W$.
Then, the routing process iteratively clusters to adjust the coupling coefficients $c_{ij}$ of each lower-level capsule $u_{l,i}$ to all higher-level capsules, 
with more crucial capsules receiving larger $c_{ij}$.  
The algorithm is in \cref{Dynamic Routing}.

\begin{figure}[t]
  \centering
  \includegraphics[scale=0.22]{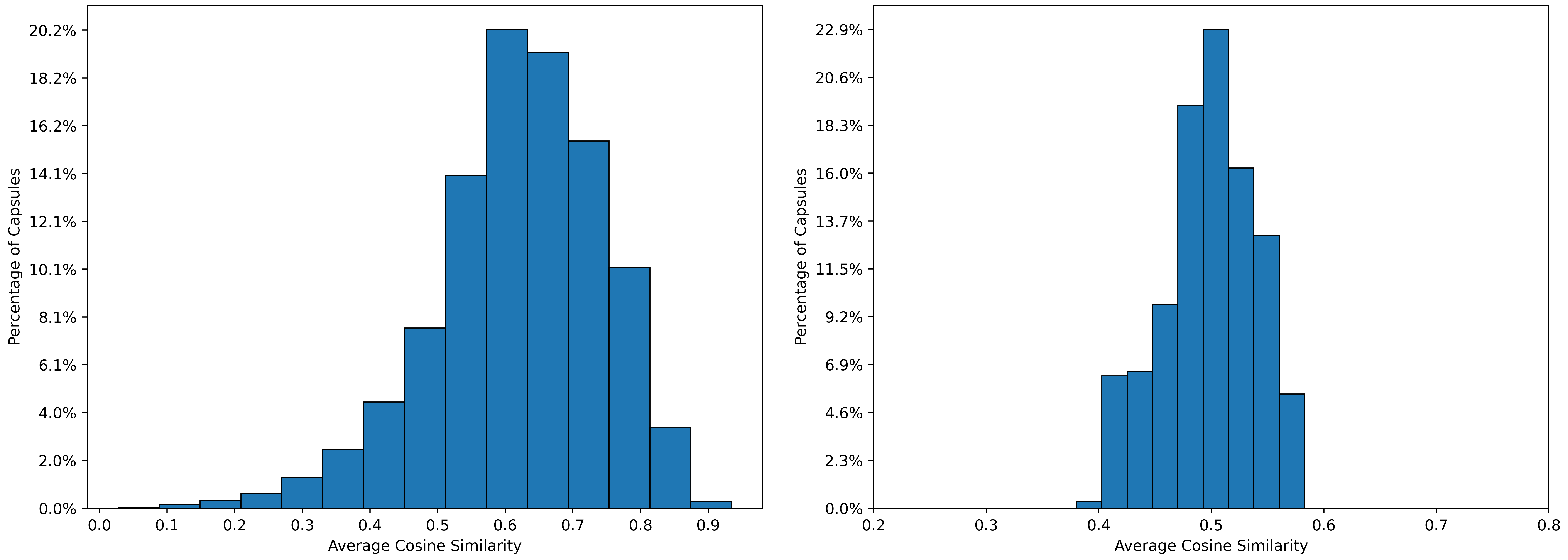}
  \caption{\textbf{Left:} In the primary capsule layer of CapsNet, 48.2\% of capsule pairs have cosine similarities greater than 0.65, indicating significant redundancy among capsules.
  \textbf{Right:} After introducing the Pruned Layer, capsule similarities effectively decrease.
        (Detailed in Section 3.2)}
  \label{similarities}
  \vspace*{-15pt}
\end{figure}

Recent studies have mentioned that Capsnet contains redundant capsules \cite{chen2022fast,sharifi2021prunedcaps,renzulli2022towards}.
As evidence, \cref{similarities} shows 48.2\% of primary capsule pairs exhibit cosine similarities above 0.65, indicating significant redundancy.
Although certain studies have implemented pruning techniques at the primary capsule layer \cite{renzulli2022rem}, 
deeper layers still show considerable over-similar issues, as demonstrated in \cref{pruned}. 
We attribute this persistent redundancy in deeper layers to dynamic routing.
On the one hand, since $v_{l+1,j} = g(\sum_i c_{ij} \hat{u}_{l,i})$, 
each higher-level capsule is essentially a weighted sum of lower-level capsules, 
indicating a fully connected structure between lower and higher layers in CapsNet \cite{jeong2019ladder}.
This full connection leads to a potential transmission of redundant information. 
On the other hand, considering $\hat{U}_l = W U_l$,
we can express higher-level capsules as
$V_{l+1} = g[(C*W)U_l]$.
Non-orthogonal matrices $C*W$ in routing 
may increase the reduced capsule similarity after pruning,
which not only impairs routing performance but also reintroduces redundancies in subsequent layers.
Additionally, dynamic routing requires multiple iterations to repeatedly update $c_{ij}$ until convergence, 
further straining computational resources.

Inspired by the successful use of orthogonality in CNNs\cite{wang2020orthogonal} and Transformer\cite{huang2022orthogonal} to reduce filter overlaps, 
we introduce the Orthogonal Capsule Network(OrthCaps). 
OrthCaps has two versions: the lightweight OrthCaps-Shallow \textbf{(OrthCaps-S)} and the OrthCaps-Deep \textbf{(OrthCaps-D)}.
OrthCaps addresses the above problems of \textbf{the fully connected structure of dynamic routing, increasing similarity in deep layers, and the need for iteration},
detailed as follows:

Firstly, we introduce a pruned capsule layer after the primary capsule layer,
which eliminates redundant capsules and retains only essential and representative ones. 
Here, capsules are firstly ordered by importance, then their cosine similarity is calculated to identify redundant capsules.  
Beginning with the least important, 
the process consistently prunes capsules that exceed the similarity threshold, proceeding through the entire set of capsules.

Secondly, to solve the iteration issue, dynamic routing is replaced with attention routing, which is a straightforward routing mechanism. 
To solve the fully connected problem, we leverage sparsemax-based attention to produce an attention map,  
which selectively amplifies relevant feature groups corresponding to specific entities while downplaying irrelevant ones.
For OrthCaps-S, a simplified attention-routing is adopted, optimizing parameter counts.

Thirdly, to address the issue of increased capsule similarity in deeper layers, we introduce orthogonality into capsule networks.
By applying Householder orthogonal decomposition, 
we enforce orthogonality in the weight matrices during attention routing.
Orthogonal weight matrices sustain low inter-capsule correlation,
which encourages fewer capsules to represent more features during backpropagation, 
thereby enhancing accuracy while effectively reducing the number of parameters.

\vspace*{5pt}
\noindent \textbf{Contributions.}
To summarize our work, we make the following contributions:

1) To our knowledge, this approach addresses the issue of deep redundancy in Capsule Networks for the first time.
A novel pruned strategy is implemented to alleviate capsule redundancy
and an orthogonal sparse attention routing mechanism is proposed to replace dynamic routing.

2) It is the first time orthogonality has been introduced into Capsule Networks as far as we know. 
This simple, penalty-free orthogonalization method is also adaptable to other neural networks.

3) Two OrthCaps versions are created: OrthCaps-S and OrthCaps-D.
OrthCaps-S sets a new benchmark in accuracy with just 1.25\% of CapsNet's parameters on datasets of MNIST, SVHN, smallNORB, and CIFAR10.
OrthCaps-D excels on CIFAR10, CIFAR100 and FashionMNIST while keeping parameters minimal.

\section{Related Work}

\noindent
\textbf{Capsule Networks.}
Dynamic routing was first introduced by Sabour et al.\cite{sabour2017dynamic}.
Though numerous studies have used attention strategies to improve dynamic routing, 
the issue of the fully connected structure and reintroduction of redundancy remains unaddressed \cite{hoogi2019self,peng2020bg,mazzia2021efficient}. 
Choi et al. incorporated attention into the capsule routing via a non-iterative feed-forward operation \cite{choi2019attention}.
Tsai et al. introduced a parallel iterative routing, which did not address the complexity of iterative requirements \cite{tsai2020capsules}. 
Furthermore, many works focused on pruning but did not mention new redundancies introduced by dynamic routing \cite{jeong2019ladder,sharifi2021prunedcaps,renzulli2022rem}. 
Jeong et al. established a ladder structure, using a pruning algorithm based on encoding \cite{jeong2019ladder}.
Sharifi et al. created a pruning layer based on Taylor Decomposition \cite{sharifi2021prunedcaps}.
Renzulli et al. used LOBSTER to create a sparse tree \cite{renzulli2022rem}.
Different from existing works, we incorporate pruning, orthogonality and sparsity to effectively reduce redundancy.

\begin{figure*}[t]
   \vspace*{-22pt}
   \centering
   \includegraphics[width=1\textwidth]{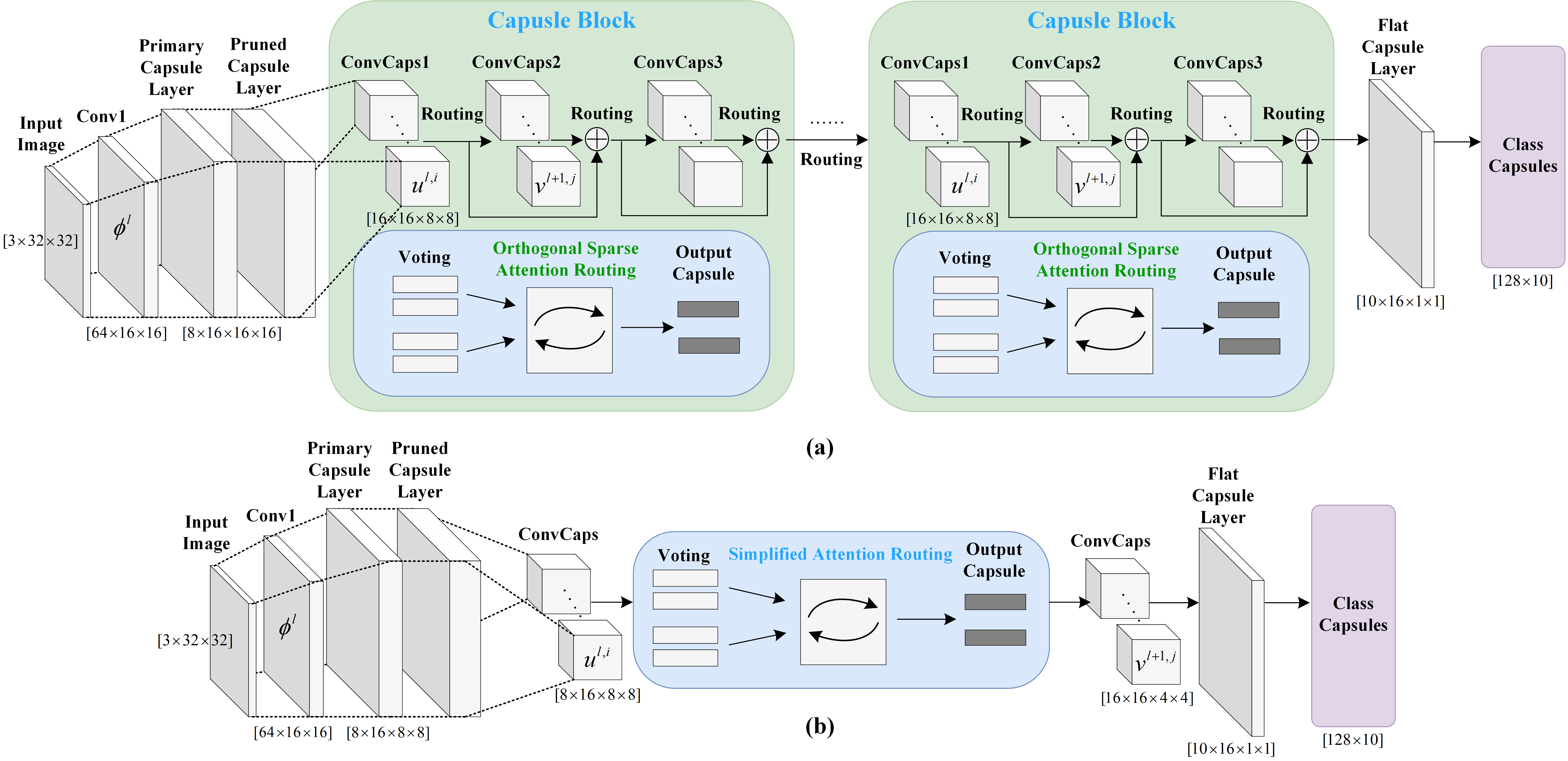}
   \caption{\textbf{(a):} In CIFAR10 classification task, the OrthCaps-D model comprises 7 capsule blocks, 
   each with 3 capsule layers, interconnected via shortcut connections and orthogonal sparse attention routing.
            \textbf{(b):} The OrthCaps-S model contains two capsule layers coping with CIFAR10 and does not use any capsule layer with MNIST. 
            These layers are linked through simplified attention routing.}
   \label{orth_structure}
   \vspace*{-10pt}
\end{figure*}

\noindent \textbf{Orthogonality.}
Various methods were proposed to introduce orthogonality into neural networks, 
which can be categorized into hard and soft orthogonality. 
Hard orthogonality maintains matrix orthogonality throughout training by either optimizing over the Stiefel manifold \cite{li2020efficient,huang2018orthogonal}, 
or parameterizing a subset of orthogonal matrices \cite{trockman2021orthogonalizing,singla2021skew,virmaux2018lipschitz}.
These methods incur computational overhead and result in vanishing or exploding gradients.
Soft orthogonality, on the other hand, employs a regularization term in the loss function 
to encourage orthogonality among column vectors of weight matrix without strict enforcement \cite{wang2020orthogonal,qi2020deep,huang2020controllable}. 
Yet, strong regularization overshadows the primary task loss, while weak regularization fails to encourage orthogonality. 
We leverage Householder orthogonal decomposition \cite{uhlig2001constructive,mathiasen2020if} to achieve strict matrix orthogonality, 
minimizing computational complexity and obviating the need for additional regularization terms.


\section{Methodology}
\label{3}


\subsection{Overall Architecture}
We introduce OrthCaps, offering both shallow (OrthCaps-S) and deep (OrthCaps-D) architectures 
to minimize parameter counts while exploring the potential for deep multi-layer capsule networks. 

As illustrated in \cref{orth_structure}a,
OrthCaps-D comprises five key components: a convolutional layer, a primary capsule layer, a pruned capsule layer, seven capsule blocks and a flat capsule layer.
Given input images \( x \in \mathbb{R}^{(B,3,W,H)} \), 
initial features \( \Phi^0 \in \mathbb{R}^{(B,C,W^0,H^0)} \) are extracted through a convolutional layer. 
The primary capsule layer generates initial capsules \( u^1 \in \mathbb{R}^{(B,n,d,W^1,H^1)} \) 
with a kernel size of 3 and stride of 2.
$B,n,d,C$ represent the batch size, number of capsules, capsule dimensions and channels, respectively.  
A pruned capsule layer is then placed to remove redundant capsules.
OrthCaps-D has seven capsule blocks, 
each containing three depthwise convolutional capsule layers(ConvCaps Layers) linked by shortcut connections to prevent vanishing gradient.
Within each block, lower-level capsules \( u^l \) are routed to the next layer \( v^{l+1} \) via orthogonal sparse attention routing.
Blocks are also connected through routing, 
allowing for stacking to construct deeper capsule networks.
The flatcaps layer is employed to map capsules into classification categories for final classification tasks.

OrthCaps-S, as illustrated in \cref{orth_structure}b, 
replaces the complete attention routing with a simplified version 
and has a single block within capsule layers.
The number of layers can be adjusted as needed.
Convolutional capsules in the primary layer utilize a 9x9 kernel with a stride of 1,
and other layers are consistent with OrthCaps-D.

\subsection{Pruned Capsule Layer}
\label{Pruned Capsule Layer}

\begin{figure*}[t]
  \vspace*{-20pt}
  \begin{center}
  \includegraphics[width=0.9\textwidth]{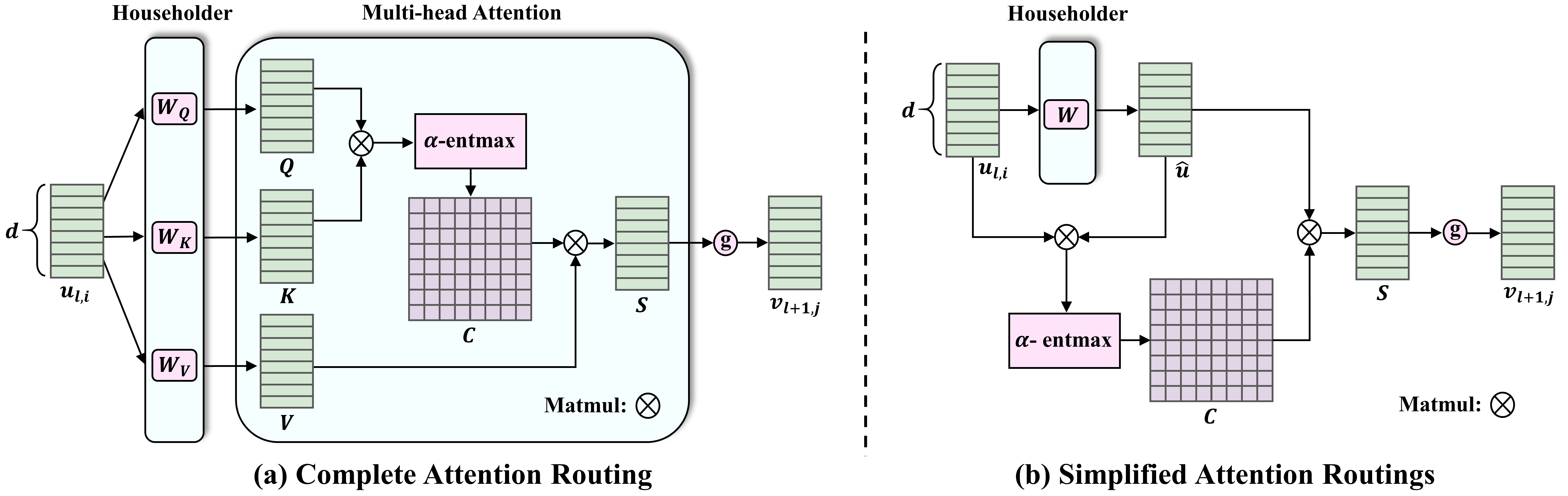}
  \end{center}
  \vspace*{-15pt}
  \caption{Orthogonal self-attention routing.}
  \label{self_attention}
  \vspace*{-15pt}
\end{figure*}


The generation of capsules starts with the primary capsule layer.  
At this initial stage, it is crucial to generate low-correlated capsules,  
which ensures efficient feature representation and reduces feature overlap during subsequent layers.
Therefore, we introduce an efficient capsule pruning algorithm \ref{algo:efficient_pruning},
including the following parts:

\vspace*{2pt}
\noindent \textbf{Redundancy Definition.}
   Redundancy occurs when two capsules capture identical or similar features.
   Given that the direction of each capsule vector encodes specific features, 
   capsules with closer directions(or angles) indicate that they capture similar features and entities. 
   Thus, we employ the cosine similarity of capsule angles to measure redundancy.

\vspace*{2pt}
\noindent \textbf{Capsule Importance Ordering.}
For redundant capsule pairs, 
random pruning may result in losing capsules vital for accurate classification.
To ensure that the less crucial capsule is pruned first when the similarity between a pair is high,
capsules are sorted in an order based on \( \| u_{flat} \|_2 \).
We employ $L_2$-norm as it calculates the length of capsule vectors, indicating the existence probability of encoded entities,
which shows the activeness of capsules \cite{jeong2019ladder}.

\vspace*{2pt}
\noindent \textbf{Pruning.} 
After ordering, a mask matrix \( M \in \mathbb{R}^{ (1,n,1)} \) is initialized to all-ones. 
Starting with the least active capsule,
the process computes the cosine similarity 
between less active capsule $u_{\text{ordered},i}$ with more active capsule $u_{\text{ordered},j}$.
When the similarity exceeds the threshold \( \theta \), 
the corresponding column in the mask for $u_{\text{ordered},i}$ is set to 0,
indicating that less active capsule is pruned.
In this way, only the active capsules are retained all along.
The final step is applying \( M \) to $u_{\text{ordered}}$, producing pruned capsules $u_{\text{pruned}}$.
$n'$ is the number of remaining capsules after pruning. 

\vspace*{-5pt}
\begin{algorithm}
  \centering
  \caption{Efficient Capsule Pruning}
  \begin{algorithmic}[1]
      \Require Capsules $u \in \mathbb{R}^{(B, n, d, W, H)}$, threshold                   $\theta$
      \Ensure $u_{\text{pruned}} \in \mathbb{R}^{(B, n, d, W, H)}$
      \State Reshape $u$ $\rightarrow$ $u_{\text{flat}} \in \mathbb{R}^{(B, n, (d \times W \times H))}$
      \State Compute $L_2$-norm: $\| u_{\text{flat}} \|_2$
      \State Order capsules by $L_2$-norm:  $u_{\text{flat}} \rightarrow u_{\text{ordered}}$
      \State Initialize $M$: all-ones matrix
      \For{$i < j$} \\
      \hspace*{0.5em} $t_{ij} = \text{cosine\_similarity}(u_{\text{ordered},i}, u_{\text{ordered},j})$ \\
      \hspace*{0.5em} $m_{i}=0$ where $t_{ij} > \theta$
      \EndFor
      \State Prune using $M$: $u_{\text{pruned}} = u_{\text{ordered}} \odot M$
      \State Reshape $u_{\text{pruned}}$ to $u_{\text{pruned}} \in \mathbb{R}^{(B, n', d, W, H)}$
      \State \Return $u_{\text{pruned}}$
  \end{algorithmic}
  \label{algo:efficient_pruning}
\end{algorithm}
\vspace*{-7pt}

Notably, we compute the cosine similarity matrix \( T \) by broadcasting, 
which avoids explicit for-loop iteration and reduces computational complexity.
The 5D capsule tensor of $u_l$ with dimensions $d$ and number $n$,
\( [B, n, d, W, H] \),
is reshaped to
\( [B, n, d\times W\times H] \) to suit for broadcasting.


\subsection{Routing Algorithm}
\label{Routing Algorithm}

We introduce the orthogonal sparse attention routing to replace dynamic routing.
This approach eliminates the need for iteration 
and leverages sparsity to reduce redundant feature transmission.


Let \( u_{l,i} \) and \( v_{l+1,j} \) represent capsules at layer \( l \) and \( l+1 \) respectively,
each with dimension \( d \). 
We employ three weight matrices \( W_Q \), \( W_K \), \( W_V \in \mathbb{R}^{d \times d} \)
to derive queries $Q$, keys $K$, and values $V$ from \( u_{l,i} \),
$Q = W_Q \times u_{l,i},  K = W_K \times u_{l,i},  V = W_V \times u_{l,i}$.
Specifically, \( W_Q \), \( W_K \), and \( W_V \) are designed as orthogonal matrices,
enabling them to project \( u_{l, i} \) into a \( d \)-dimensional orthogonal subspace. 

As shown in \cref{self_attention}a, attention routing aims to produce coupling coefficient \( c_{ij} \), 
which serves as the weight during routing from lower-level to higher-level capsules.
The coupling coefficient matrix \( C \) is derived from the attention map, generated through the dot product of $Q$ and $K$, 
$C = \alpha\text{-Entmax}(QK^T/\sqrt{d})$.
Here, we replace the softmax of the original attention mechanism with \(\alpha\)-Entmax in \cref{alpha_entmax},
enhancing the sparsity of the attention map.
\(\alpha\)-Entmax adaptively sets small \( c_{ij} \) to zero,
thereby encouraging routing to prioritize more important capsules while minimizing irrelevant information transfer.
\vspace*{-5pt}
\begin{equation}
  \alpha\text{-Entmax}(x)_i = \max \left( \frac{x_i - \tau}{\alpha}, 0 \right)^{\frac{1}{\alpha - 1}}
  \label{alpha_entmax}
  \vspace*{-5pt}
\end{equation}
$\tau$ is a self-adaption threshold and $\alpha$ is a hyperparameter controlling the sparsity of the attention map.

The vote $s_{l+1,j}$ is computed as the product of \( V \) and $C$.
In \cref{S_{i,j}},
higher-level capsule \( v_{l+1,j} \) is generated by $s_{l+1,j}$ from a multi-head self-attention mechanism with 16 heads, using the nonlinear activation function $g$.
\vspace*{-7pt}
\begin{equation}
   v_{l+1,j} = g(s_{l+1,j}) = g(\alpha\text{-Entmax}(QK^T/\sqrt{d}) \times V)
   \label{S_{i,j}}
    \vspace*{-5pt}
\end{equation}


For simplified attention-routing in \cref{self_attention}b, 
we condense prediction matrices \( W \) from three to one
and replace \( K, Q, V \) with \( u_{l, i} \) \cite{hinton2023represent}. 
The \( \hat u_{l,i} \) is the prediction for \( v_{l+1,j} \). 
The attention map $C$ is obtained using \( \alpha \)-entmax with the dot product to produce the vote \( s_{l+1,j} \) 
$= \hat u_{l, i} \times C = \hat u_{l, i} \times (\alpha\text{-Entmax}(\hat u_{l, i} u_{l, i}^T/\sqrt{d}))$. 
\( s_{l+1,j} \) is processed through \( g \) to produce \( v_{l+1,j} \).
Notably, standard convolutions are supplanted by depthwise convolutions to minimize parameter count.
Without any iteration, attention routing reduces computational complexity.

\vspace*{-2pt}
\subsection{Orthogonalization}
\label{Orthogonalization}

In \cref{Pruned Capsule Layer}, we reduce redundancy by pruning highly similar capsules.
To preserve pruning effect in subsequent layers,
it's vital to maintain low similarity among capsules all along.  
Here, capsules with diverse angles span a broader multi-dimensional feature subspace, 
enabling the network to capture a wider array of features with fewer capsules, 
which boosts accuracy and reduces parameter count. 

We utilize orthogonalization to achieve this. 
An orthogonal matrix, signifying a rotation or reflection transformation, 
keeps vector lengths and inter-vector angles constant during multiplication with a vector set.
Considering that cosine similarity quantifies angles between vectors, applying an orthogonal matrix 
$W$ to a vector set retains the mutual cosine similarity among all vector pairs in the set. 
Let \( \{v_{l+1,j} \mid j = 1, 2, \ldots, m\} \) be a set of capsule vectors at layer \( l+1 \),
we derive Lemma 1: 

\noindent \textbf{Lemma 1}:
For any \( i, j \in \{1, 2, \ldots, m\} \), if \( W \) is orthogonal, 
the cosine similarity between \( v_{l+1,i} \) and \( v_{l+1,j} \) remains unchanged after multiplication by \( W \).

\cref{Orthogonalization of Weight Matrices} discusses the selection of targets for orthogonalization, 
while \cref{Householder Orthogonalization} details the method.
 

\vspace*{-7pt}
\subsubsection{Orthogonalization of Weight Matrices}
\label{Orthogonalization of Weight Matrices}

The goal of orthogonalization is to maintain low similarity among higher-level capsules.
Following \cref{Routing Algorithm}, we represent higher-level capsules into matrix multiplication:
\vspace*{-7pt}
\begin{equation}
  V_{l+1} = g(S_l) = g(CV) = g[(C \times W_V) U_l]
  \label{Voting}
  \vspace*{-5pt}
\end{equation}

As the activation function $g$ in CapsNet preserves the capsule vector's direction \cite{sabour2017dynamic}, 
in line with Lemma 1, ensuring orthogonality of $C$ and $W_V$ can maintain low similarity.
However, orthogonalizing $C$ directly is hard,
so we delve deeper into its calculation process in \cref{C}:
\vspace*{-5pt}
\begin{equation}
  \scalebox{0.8}{
  $C = \alpha\text{-Entmax}(QK^T/\sqrt{d}) = \alpha\text{-Entmax}(W_{Q}U_{l}{U_l}^{T}{W_K}^T/\sqrt{d})$ 
  }
  \label{C}
  \vspace*{-5pt}
\end{equation}

Since $d$ is a constant, it requires no additional analysis.
The impact of $U_{l}{U_l}^{T}$ on orthogonalizing $C$ is mitigated by pruning,
which removes capsules with short lengths and high cosine similarity. 
Consequently, the remaining capsules were updated to approximate unit length and low correlation, 
akin to a standard orthonormal basis.
Thus, $U_{l}{U_l}^{T}$ gradually becomes more orthogonal as the network trains, 
minimizing its impact on the orthogonality of $C$. 
Although $\alpha\text{-Entmax}$, a nonlinear function, may not preserve the orthogonality of inputs,
it renders $C$ sparse. 
This sparsity directs lower-level capsules effectively toward their respective higher-level targets, 
which reduces interference during routing, 
thus encouraging a relatively low similarity among higher-level capsules.

Our above analysis indicates that orthogonalizing $W_Q, W_K$, and $W_V$ is essential for maintaining low cosine similarity among higher-level capsules. 
While $C$ is not fully orthogonalized, our experiments demonstrate considerable improvements in both accuracy and parameter efficiency.

\begin{figure}[t]
  \vspace*{-10pt}
  \begin{center}
  \includegraphics[width=0.45\textwidth]{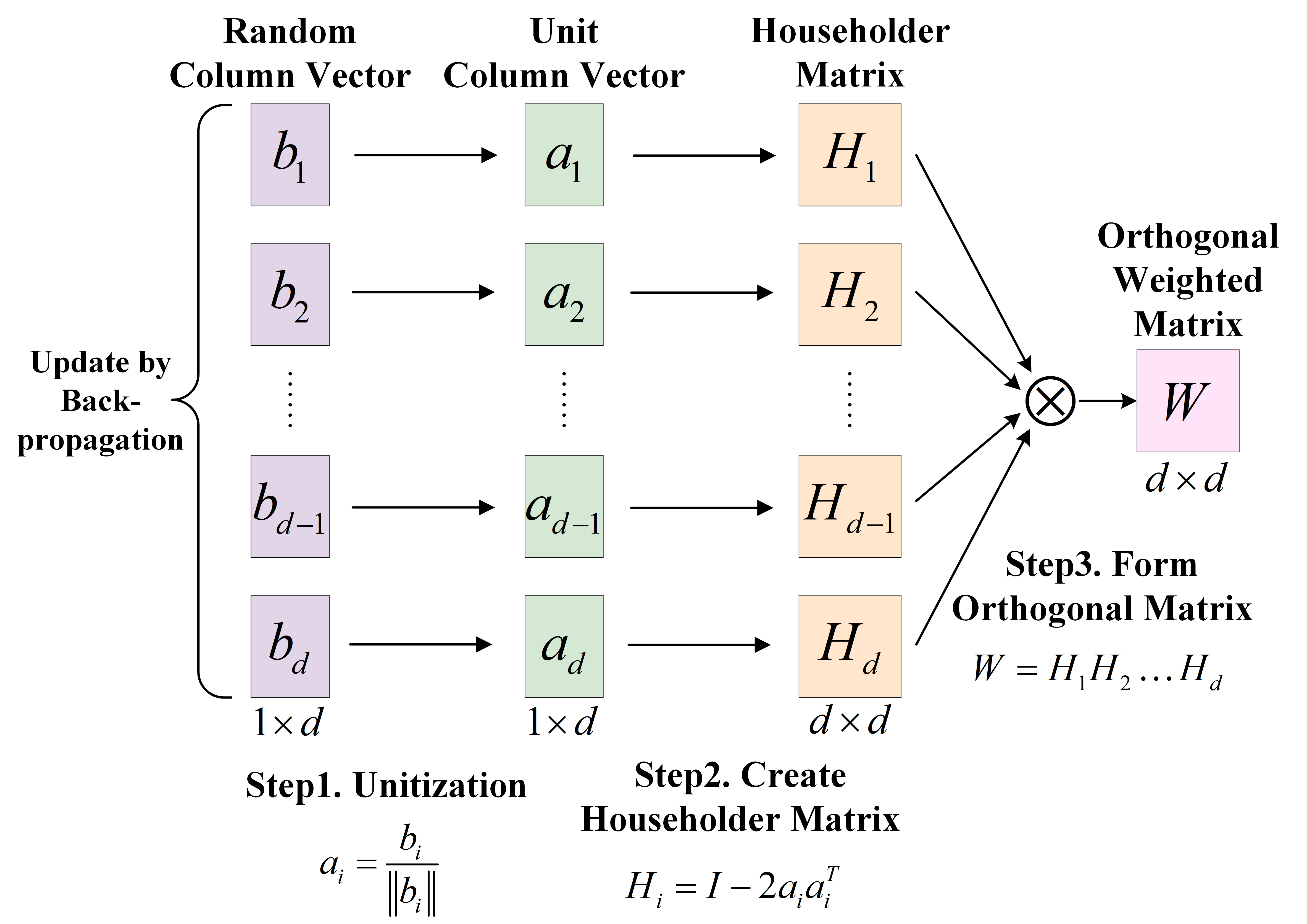}
  \end{center}
  \vspace*{-15pt}
  \caption{The computing process of HouseHolder method.}
  \label{Householder}
  \vspace*{-15pt}
\end{figure}

\vspace*{-10pt}
\subsubsection{Householder Orthogonalization}
\label{Householder Orthogonalization}

Let \( W \) be the weight matrix requiring orthogonalization. 
As shown in \cref{Householder}, the Householder orthogonal decomposition theorem is employed to formulate an endogenously optimizable orthogonal matrix. 
The essence of this approach is in the following algebraic lemma \cite{uhlig2001constructive}:

\noindent \textbf{Lemma 2}: Any orthogonal \(n \times n\) matrix is the product 
of at most \(n\) orthogonal Householder transformations.

Based on Lemma 2, an orthogonal matrix \(W \in \mathbb{R}^{d \times d}\) can be formulated in \cref{W1}:
\vspace*{-5pt}
\begin{equation}
   W = H_0 H_1 \dots H_{d-1}
   \label{W1}
   \vspace*{-5pt}
\end{equation}

Each \(H_i\) represents a Householder transformation, defined as \(H_i = I - 2a_i a_i^T\), 
where \(a_i\) is a unit column vector. 
We utilize a set of randomly generated column vectors 
\(\{b_i | i = 0, \dots, d - 1\}\) instead of \(a_i\) to construct \(H_i\) as detailed in \cref{W2}.
During training,  \(b_i\) is optimized through gradient backpropagation.
\(W\) inherently preserves its orthogonality during training.
\vspace*{-5pt}
\begin{equation}
   W = \prod_{i=0}^{d-1} \left( I - \frac{2 b_i b_i^T}{{\|b_i\|}^2} \right)
   \label{W2}
   \vspace*{-5pt}
\end{equation}

\noindent \textbf{Lemma 3}: \( W_Q \), \( W_K \), and \( W_V \) constructed using Equation \cref{W2} are Orthogonal.

Following \cref{W2}, 
\( W_Q \), \( W_K \), and \( W_V \) could easily be orthogonalized, 
where the proof is provided in supplement material \cref{HouseHolder Proof}.
Householder orthogonalization enables computationally efficient transformation of 
arbitrary coefficient matrices into orthogonal matrices without any additional penalty terms in the loss function.


\section{Experiments}

\subsection{Experimental Setup}

\begin{table*}
    \vspace*{-15pt}
    \centering
    \caption{\textbf{(a): } OrthCaps-S ranks as the top or second best across five datasets, 
                    standing out as being resource-efficient with only 105.5K parameters and 673.1M FLOPS.
                    \textbf{(b): } OrthCaps-D shows competitive performance with fewer parameters and
                    less computational cost. }
    \label{table1}
    \vspace*{-5pt}

    \resizebox{0.75\linewidth}{!}{
       \begin{tabular}{cccccccc}
            \toprule
            Shallow Networks & Param$\downarrow$ & FLOPS[M]$\downarrow$ & MNIST  & SVHN & smallNORB & CIFAR10 \\ 
            \midrule
            OrthCaps-S & \textbf{105.5K} & 673.1 & \textbf{99.68} & \textbf{96.26} & \textbf{98.30} & \textbf{86.84} \\
            Efficient-Caps & 162.4K & \textbf{631.1} & 99.58  & 93.12 & 97.46 & 81.51 \\
            CapsNet & 8388K & 803.8 & 99.52 & 91.36 & 95.42 & 68.72 \\
            Matrix-CapsNet with EM routing & 450K & 949.6 & 99.56 & 87.42 & 95.56 & 81.39 \\
            AR CapsNet & 9.1M & 2562.7 & 99.46 & 85.98 & 96.47 & 85.39 \\
            DA-CapsNet & 7M* & - & 99.53* & 94.82* & 98.26* & 85.47* \\
            AA-CapsNet & 6.6M* & - & 99.34* & 91.23* & 89.72* & 79.41* \\
            \midrule
            Baseline CNN & 4.6M & 1326.9 & 99.22 & 91.28 & 87.11 & 72.20 \\
            \bottomrule
       \end{tabular}
    }
    \subcaption*{(a)}

    \vspace*{5pt}

    \resizebox{0.75\linewidth}{!}{
       \begin{tabular}{ccccccc}
            \toprule
            Deep Networks & Param $\downarrow$ & FLOPS[M]$\downarrow$ & CIFAR10 & CIFAR100 & MNIST & FashionMNIST \\ 
            \midrule
            OrthCaps-D & \textbf{574K} & 3345 & 90.56 & \textbf{70.56} & \textbf{99.59} & \textbf{94.60} \\
            AR-CapsNet(7 ensembled) & 6.3M & 16657.5 & 88.94 & 56.53 & 99.49 & 91.73 \\
            CapsNet(7 ensembled) & 5.8M* & 5137.4* & 89.4* & - & - & - \\
            Inverted Dot-Product & 1.4M & 5340.9 & 84.98 & 57.32 & 99.35 & 92.85 \\
            RS-CapsNet & 5.0M* & - & 89.81* & 64.14* & - & 93.51* \\
            DeepCaps & 13.5M & \textbf{2687} & \textbf{91.01} & 69.72 & 99.46 & 92.52 \\
            \midrule
            ResNet-18\footnotemark[1] & 11.7M & 5578.8 & 95.10 & 77.60 & 99.29 & 93.32  \\
            VGG-16\footnotemark[1] & 147.3M & 15143.1 & 93.57 & 73.10 & 99.21 & 92.21 \\
            \bottomrule
       \end{tabular}
    }
    \subcaption*{(b)}
    
    \vspace*{-10pt}
\end{table*}

\noindent \textbf{Implementation Details and Datasets}
\vspace*{5pt}

OrthCaps was developed using PyTorch 1.12.1, running on Python 3.9, and the training process was accelerated using four GTX-3090 GPUs.
We adopted the margin loss as defined in \cite{sabour2017dynamic}. 
Observing minimal performance benefits in our experiments, we decided to exclude the reconstruction loss. 
Our model utilized the AdamW optimizer with a cosine annealing learning rate scheduler and a 5-cycle linear warm-up. 
We set learning rate at 5e-3, weight decay at 5e-4, and batchsize at 512. 
We conducted experiments on SVHN \cite{netzer2011reading}, smallNORB \cite{lecun2004learning}, CIFAR10, and MNIST \cite{lecun1998mnist} for OrthCaps-S.
OrthCaps-D was evaluated on CIFAR10, CIFAR100 \cite{krizhevsky2009learning}, Fashion-MNIST \cite{xiao2017fashion}, and MNIST.
We resized SmallNORB from $96\times96$ to $64\times64$ and cropped it to $48\times48$ like \cite{sabour2017dynamic}. All other datasets retained original sizes.
For data augmentation, we adopted the methods outlined in \cite{hinton2018matrix}.
For reproducibility, we detailed hyperparameters and setups in supplement material \cref{hyperparameters}.


\setcounter{footnote}{1}
\footnotetext{\url{https://github.com/kuangliu/pytorch-cifar}}

\vspace*{5pt}
\noindent \textbf{Comparison Baselines}
\vspace*{5pt}

OrthCaps was benchmarked against a range of baseline models to evaluate its performance. 
For OrthCaps-S, 
we compared it with Efficient-Caps\cite{mazzia2021efficient}, CapsNet\cite{sabour2017dynamic}, Matrix-CapsNet with EM routing\cite{hinton2018matrix}, 
AR-CapsNet\cite{choi2019attention}, AA-CapsNet\cite{pucci2021self}, DA-CapsNet\cite{huang2020capsnet} and standard 7-layer CNN. 
For OrthCaps-D, we used baselines such as 
CapsNet (7 ensembles), AR-CapsNet (7 ensembles), RS-CapsNet\cite{9086631}, 
Inverted Dot-Product\cite{tsai2020capsules}, DeepCaps\cite{rajasegaran2019deepcaps}, ResNet-18\cite{he2016deep}, and VGG-16\cite{simonyan2014very}. 
All comparative results were derived from running official codes with our hyperparameters.

\vspace*{-3pt}
\subsection{Classification Performance Comparison}

\cref{table1} illustrates the classification performance of OrthCaps-S and OrthCaps-D,
with model sizes denoted by Param and computational demands represented as FLOPS[M]. 
All models utilize a backbone of 4 convolutional layers and undergo training for 300 epochs.
The Param and FLOPS[M] of each table are tested on MNIST and CIFAR10, respectively.  
An asterisk (*) signifies that no official code is available, so we refer to the model performance stated in the original papers. 

As shown in \cref{table1}a, 
OrthCaps-S achieves superior efficiency with merely 105.5K parameters, outperforming CNN, CapsNet, and many variants. 
For instance, Efficient-Caps, a state-of-the-art model on efficiency, has about 50\% more parameters.
Despite its compact design, OrthCaps-S either outperforms or matches the performance of other capsule networks across all four datasets. 
On the SVHN and CIFAR10, OrthCaps-S achieves accuracies of 96.26\% and 87.92\%, respectively, surpassing CapsNet which has 80 times more parameters.
With a computational demand of 673.1M FLOPS, 
it's worth noting that the slight increase in FLOPS compared to Efficient-Caps 
is due to the additional computations from the pruned capsule layer and orthogonal transformations. 
Given the substantial decrease in parameter count and the enhanced accuracy, this FLOPS trade-off is warranted.

For OrthCaps-D, as illustrated in \cref{table1}b, 
it exhibits competitive performance with fewer parameters and less computational cost on complex datasets.
Although convolution-based networks such as ResNet-18 and VGG-16 perform well on CIFAR10 and CIFAR100, 
OrthCaps-D offers competitive performance using just 1.41\% and 0.11\% of their parameters as well as 56\% and 20.8\% of their FLOPS, respectively.
The efficiency of OrthCaps becomes evident when compared with DeepCaps.
While DeepCaps achieves a 91.01\% accuracy on CIFAR10, 
its significant parameter count of 13.42M highlights a compromise. 
It's noteworthy that both OrthCaps variants maintain high performance with fewer parameters.

\subsection{Ablation Study}
 
\subsubsection{Orthogonal Attention Routing}

Through a cross-comparison of frames-per-second (FPS) and accuracy on two datasets with different complexity, 
as shown in \cref{routing}, 
we compare attention routing with dynamic routing\cite{sabour2017dynamic} and sparse softmax with standard softmax, respectively.
Additionally, $\alpha$ is settled to 1.5 in our experiments according to \cite{peters2019sparse}. 

\begin{table}[t]
    \vspace*{-10pt}
    \centering
    \caption{Comparison of Orthogonal sparse attention routing and dynamic routing algorithms. 
    FPS is tested under MNIST dataset.}
    \vspace*{-5pt}
    \resizebox{1\linewidth}{!}{
       \begin{tabular}{cccc}
          \toprule
          Variants & FPS$\uparrow$ & MNIST & CIFAR10 \\ 
          \midrule
          Attention routing \& $\alpha$-entmax \& orthogonality & 1639 & 99.68 & 86.92 \\
          Attention routing \& softmax & 1785 & 99.62 & 83.44   \\
          Dynamic routing \& $\alpha$-entmax \& orthogonality & 1232 & 99.51 & 70.01 \\
          Dynamic routing \& softmax & 1339 & 99.49 & 68.72\\
          \bottomrule
       \end{tabular}
    }
    \label{routing}
    \vspace*{-10pt}
\end{table}

Attention routing consistently outperforms dynamic routing in both classification accuracy and processing speed, 
achieving a 25.8\% speed enhancement on average. 
Even with a faster softmax, 
dynamic routing only reaches 1339 FPS, indicating its inherent computational inefficiencies.  
Although $\alpha$-entmax's complexity and the additional computational demands from orthogonality slightly reduce processing speed compared to softmax, 
this trade-off is justified by a substantial increase in accuracy and robustness (detailed in \cref{robustness}).
Overall, our attention routing combined with $\alpha$-entmax and orthogonality 
balances performance and computational efficiency.

\vspace*{-10pt}
\subsubsection{Pruned Capsule Layer}

\cref{similarities} illustrates that by integrating the pruned layer, 
the average capsule similarity decreases due to redundant capsule elimination.  
Consequently, as the capsule count reduces, the dimensions of the associated prediction matrix diminish, 
thereby lowering the parameter count. 
This is proved in \cref{pruning},
where the pruned OrthCaps-S reduces parameters from 127K to 105K without sacrificing performance. 
In fact, accuracy improves from 99.53\% to 99.68\% and from 85.32\% to 87.92\% on MNIST and CIFAR10, respectively.  
Similarly, applying pruning to CapsNet results in higher accuracy with reduced parameters (7492K from 8388K).
This shows our pruning method's efficacy in streamlining the model and enhancing performance.

\begin{table}[t]
    \vspace*{-10pt}
    \centering
    \caption{CapsNets are compared with and without the pruning layer, with the similarity threshold set to 0.7. 
    Param[K] is tested on MNIST.}
    \vspace*{-5pt}
    \resizebox{0.8\linewidth}{!}{
       \begin{tabular}{cccc}
             \toprule
             Variant & Param[K]$\downarrow$ & MNIST & CIFAR10 \\ 
             \midrule
             OrthCaps-S with pruning & \textbf{105} & \textbf{99.68} & \textbf{87.92} \\
             OrthCaps-S & 157 & 99.53 & 85.32 \\
             Capsnet with pruning & 7492 & 99.51 & 71.08 \\
             Capsnet & 8388 & 99.42 & 68.72 \\
             \bottomrule
       \end{tabular}
    }
    \label{pruning}
    \vspace*{-5pt}
\end{table}

\begin{figure}[t]
    \centering
    \includegraphics[scale=0.35]{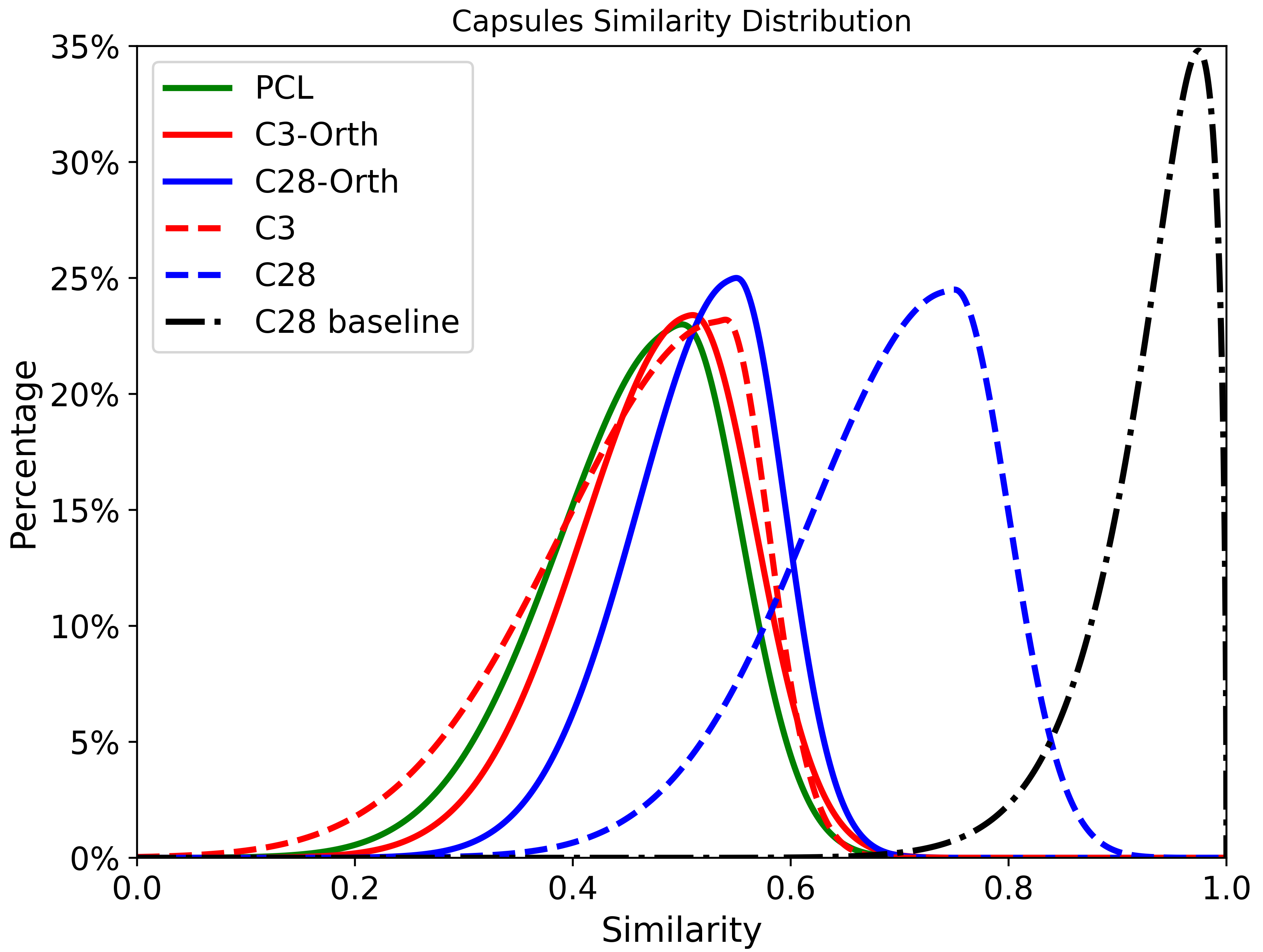}
    \vspace*{-5pt}
    \caption{Redundancy comparison between different pruning strategies. \textbf{(The more to the left, the better.)}
    The x-axis shows capsule similarity; 
    the y-axis indicates capsule count percentage. 
    PCL, C3, and C28 mark the primary, third, and twenty-eighth capsule layers. 
    Solid lines (C3-Orth, C28-Orth) and dashed lines represent pruning with and without orthogonality, respectively; 
    the dash-dot line indicates the absence of both pruning and orthogonality.
    Tests are on OrthCaps-D with CIFAR10 dataset.}
    \label{pruned}
    \vspace*{-15pt}
\end{figure}

\cref{pruned} illustrates the necessity of incorporating pruning with orthogonality. 
Capsule similarity is gauged with cosine similarity to measure the redundancy as mentioned above. 
As the network goes deeper, the dashed line (indicating pruning without orthogonality) shifts rightward, 
suggesting an increase in capsule similarity.
This shift proves that non-orthogonal weight matrices reintroduce redundancy. 
However, the solid line (indicating pruning with orthogonality) shows consistently low capsule similarity.
Even at 28 layers deep, the similarity remains low, 
affirming the efficacy of orthogonality in preserving capsule directions to maintain low inter-capsule correlations.
The black dash-dot line denotes similarity without orthogonality and pruning, 
exhibiting the highest redundancy, further evaluating the effectiveness of our method.

\subsection{Similarity Threshold}

To find the optimal similarity threshold $\theta$,
we evaluated the classification accuracy on three datasets and capsule number after pruning (N) for thresholds ranging from 0.3 to 1.0.
At $\theta = 1.0$, the pruning layer becomes ineffective as it targets capsules with similarities above one.

As \cref{threshold} shows, the optimal accuracy occurs at $\theta = 0.7$. 
A threshold below 0.7 leads to excessive pruning, 
over-reducing capsule numbers and causing feature loss, thus impacting the accuracy. 
Notably, at $\theta = 0.4$, all three datasets show a marked accuracy drop, 
indicating a critical point where key information is lost.
CIFAR10 is the most affected, likely due to its complex background and rich features,
making it more sensitive to excessive pruning. 
Conversely, a higher $\theta$ weakens pruning effectiveness.
As redundant information accumulates, the classification capsules become disrupted, 
slightly diminishing performance. 
With overall consideration, we set $\theta$ into 0.7.

\begin{table}
    \vspace*{-10pt}
    \centering
    \caption{Comparison of different similarity thresholds on MNIST, SVHN and CIFAR10.}
    \vspace*{-5pt}
    \resizebox{0.85\linewidth}{!}{
    \begin{tabular}{ccccc}
        \toprule
        $\theta$ & N & MNIST & SVHN & CIFAR10\\ 
        \midrule
        0.3 & 58 & 38.25 & 18.96 & 10.25 \\
        0.4 & 116 & 43.60 & 24.87 & 11.03 \\
        0.5 & 303 & 90.09 & 86.12 & 63.35\\
        0.6 & 676 & 97.46 & 94.66 & 81.50 \\
        \textbf{0.7} & 952 & \textbf{99.68} & \textbf{96.26} & 87.89 \\
        0.8 & 1093 & 99.63 & 96.13 & \textbf{87.92} \\
        0.9 & 1139 & 99.61 & 95.96 & 86.41 \\
        1.0(without pruning) & 1152 & 99.53 & 95.25 & 85.32 \\
        \bottomrule
        \end{tabular}
    }
    \label{threshold}
    \vspace*{-10pt}
 \end{table}



\subsection{Robustness to Adversarial Attacks}
\label{robustness}

Capsule networks have demonstrated exceptional performance in terms of robustness \cite{hinton2018matrix}.
Considering OrthCaps as it eliminates redundant capsules to suppress low $L_2$-norm capsules, 
which we consider as noise capsules \cite{de2020introducing}.
It can enhance better robustness against small perturbations.
To evaluate this, 
we conduct a robustness comparison between OrthCaps, Capsule Networks, Orthogonal CNNs (OCNN) and 7-layer CNNs using the CIFAR10 dataset. 
We employ the Projected Gradient Descent (PGD) white-box attack method \cite{guo2019simple}, setting the maximum iteration count at 
40, step size at 0.01, and the maximum perturbation at 0.1.
We assess the robustness using three metrics: attack time (AT), model query count (QC), and accuracy after attacks (ACC). 
As shown in \cref{TABLE5}, OrthCaps outperforms in all three metrics, 
confirming its superior handling of complex spatial structures. 
Specifically, OrthCaps requires 1.72 times more queries compared to CapsNet and exhibits a 
9\% higher accuracy, evaluating its robustness.

\subsection{Orthogonality}

This experiment demonstrates the effectiveness of the HouseHolder orthogonalization method 
and its advantages over other orthogonalization methods. 
We define an orthogonality metric \( O = \| K^T K - I \| \). 
In \cref{TABLE6}, the metric \( O \) decreases from 0.02 to 0.01 during training, 
showing the effectiveness of the orthogonalization method. 

\begin{table}[t]
\vspace*{-10pt} 
   \centering
   \caption{Comparison of OrthCaps, CapsNet, OCNN and baseline CNN under PGD attack. The CIFAR10 dataset is used without any data augmentation. Our results are an average of 5 test runs. }
   \label{TABLE5}
   \vspace*{-5pt}
   \resizebox{0.75\linewidth}{!}{
   \begin{tabular}{cccc}
       \toprule
       Variants & AT(s) $\uparrow$ & QC[K] $\uparrow$ & ACC $\uparrow$\\ 
       \midrule
       OrthCaps & \textbf{345.92} & \textbf{69K} & \textbf{23.52} \\
       CapsNet & 198.93 & 48K & 14.62\\
       OCNN & 136.7 & 46K & -\\
       baseline CNN & 16.65 & 10K & 0.35\\
       \bottomrule
       \end{tabular}
   }
   \vspace*{-8pt}
\end{table}


\begin{table}[t]
    \centering
    \caption{Orthogonality of weight matrices in attention routing of SVHN dataset. $O$ decreases from 0.02 to 0.01 during training.}
    \vspace*{-5pt}
    \resizebox{0.45\linewidth}{!}{
    \begin{tabular}{ccc}
            \toprule
            EPOCH  & Acc & $O$ $\downarrow$\\ 
            \midrule
            1 &  83.75 & 0.0236 \\
            10 &  98.58 & 0.0215\\
            100 &  99.42 & 0.0153  \\
            300 & 99.56 & 0.0120 \\
            \bottomrule
            \end{tabular}
    }
    \label{TABLE6}
    \vspace*{-10pt}
\end{table}

\begin{figure}[t]
    \centering
    \includegraphics[width=0.45\textwidth]{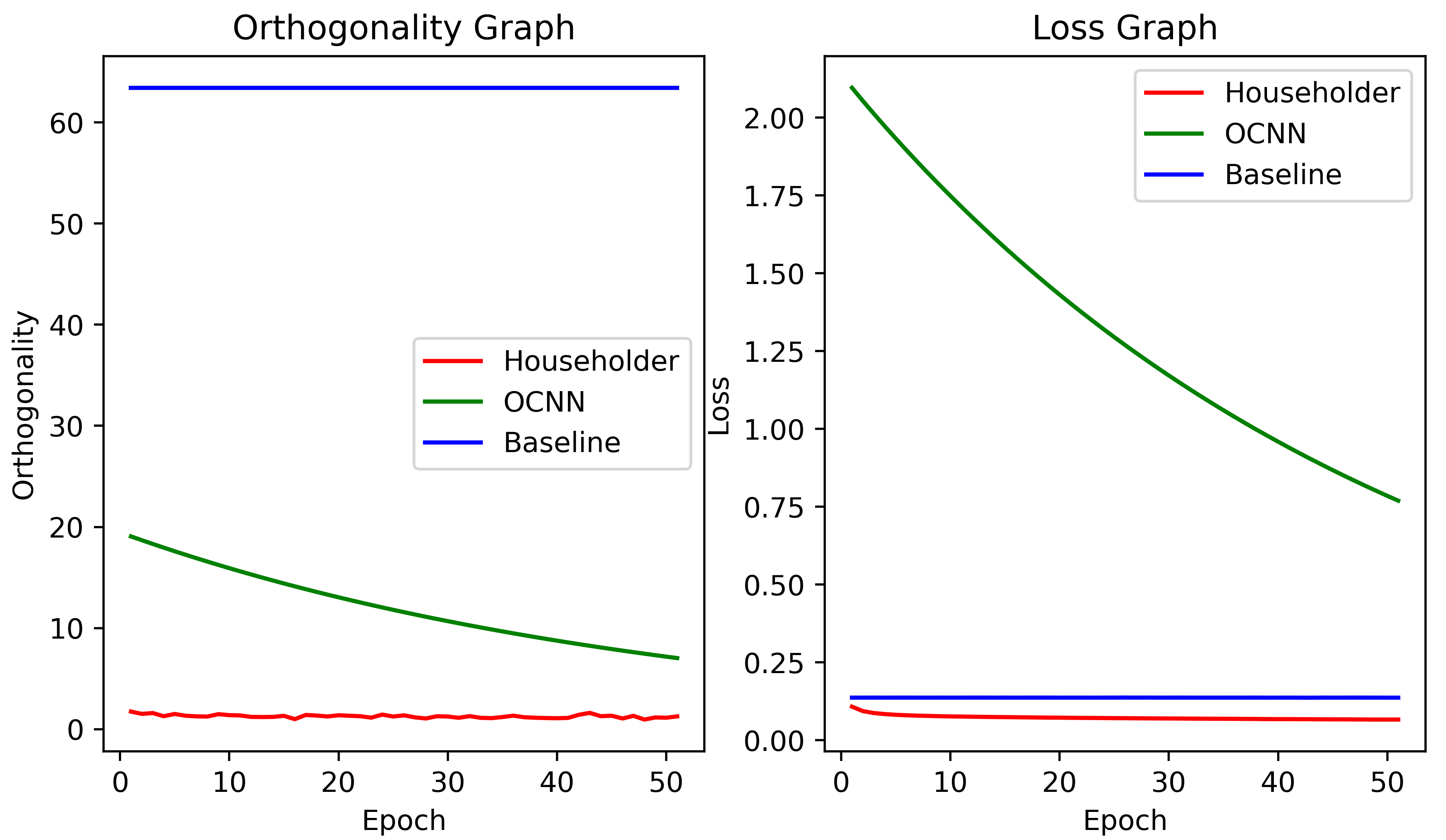}
    \vspace*{-5pt}
    \caption{Capsnet with different Orthogonal regularization on MNIST dataset. Our HouseHolder orthogonalization method reaches better orthogonality and loss decay.}
    \label{orth2}
    \vspace*{-12pt}        
 \end{figure}

We further demonstrate Householder's role as a regularization technique for neural networks.
In \cref{orth2}, our method achieves better orthogonality and loss decay than OCNN \cite{wang2020orthogonal}.
The baseline ResNet18, without any orthogonal regularization, is depicted by the blue line,
while the green and red lines stand for OCNN and our method, respectively. 
Despite the decrease in orthogonality loss for OCNN throughout training, 
it remains almost 10 times higher compared to our Householder technique. 
The near-flat trajectory of the red line testifies to Householder's consistent orthogonality preservation across the training.
Furthermore, our method registers a smaller loss than OCNN, due to its better training performance.

\vspace*{-5pt}
\section{Conclusions and Future Work}

In this study, we have introduced a novel capsule network with orthogonal sparse attention routing and pruning. 
Specifically, Householder orthogonal decomposition is used to ensure strict matrix orthogonality in attention routing without additional penalty terms, 
and the capsule pruning layer introduces sparsity into routing, minimizing capsule redundancies. 
Experiments show that OrthCaps has lower parameters and reduces computational overhead, 
overcoming the challenges of computational expense and redundancy in dynamic routing. 
On image classification tasks, OrthCaps outperforms state-of-the-art methods, 
demonstrating improved robustness. 
We look forward to further developments in this area.

\newpage

{
    \small
    \bibliographystyle{ieeenat_fullname}
    \bibliography{main}
}

\clearpage
\setcounter{page}{1}
\maketitlesupplementary

\section{Symbols and abbreviation used in this paper}
\label{Symbols}

\begin{table}[H] 
   \centering
   \resizebox{\linewidth}{!}{
   \begin{tabular}{ll}
       \toprule
       Symbol & Description  \\ 
       \midrule
       OrthCaps & Orthogonal Capsule Network  \\
       OrthCaps-S & Shallow network variant of OrthCaps \\
       OrthCaps-D & Deep network variant of OrthCaps \\
       $x$ & Input image \\
       $l$ & Layer index \\
       $\Phi^0$ & Features from initial convolutional layer\\
       $u_{l,i}/(U_l)$ & Capsules/ (matrix) at layer $l$ \\
       $v_{l+1,i}/(V_{l+1})$ & Capsules/ (matrix) at layer $l+1$ \\
       $n$ & Capsule count in layer $l$ \\
       $m$ & Capsule count in layer $l+1$ \\
       $d$ & Capsule dimension \\
       $W$ & Feature map width \\
       $H$ & Feature map height \\
       $B$ & Batch size \\
       $u_{flat}$ & Flattened capsules \\
       $u_{ordered}$ & Capsules ordered by their $L_2$-norm \\
       $u_{pruned}$ & Pruned capsules \\
       $m_i/(M)$ & Mask column/matrix for pruned capsule layer \\
       $t_{ij}/(T)$ & Cosine similarities/ (matrix) for pruned capsule layer \\
       $\theta$ & Threshold for pruned capsule layer \\
       $Q, K, V$ & Attention routing components: Query, Key, Value \\
       $W_Q, W_K, W_V$ & Weight matrices for Q, K, V \\
       $c_{i,j}/(C)$ & Coupling coefficients/ (matrix) for attention routing \\
       $s_{l+1,j}/(S_l)$ & Votes/ (matrix) for attention routing \\
       $W$ & Weight matrix for simplified attention routing \\
       $g$ & Activation function \\
       $H$ & Householder matrix \\
       $a_{i}$ & Unit vector in Householder matrix formulation \\
       $b_{i}$ & Learnable vector in Householder matrix \\
       \bottomrule
       \end{tabular}
   }
\end{table}

\section{experimental Setups}

\subsection{hyperparameters}
\label{hyperparameters}

\begin{table}[H] 
   \centering
   \resizebox{\linewidth}{!}{
   \begin{tabular}{ll}
       \toprule
       Hyperparameter & Value  \\ 
       \midrule
       Batchsize & 512 (4 parallelled)  \\
       Learning rate & 5e-3 \\
       Weight decay & 5e-4 \\
       Optimizer & AdamW \\
       Scheduler & CosineAnnealingLR and 5-cycle linear warm-up \\
       Epochs & 300 \\
       Data augmentation & RandomHorizonFlip, RandonClip with padding of 4 \\
       Dropout & 0.25 \\
       $m^+$ & 0.9 \\
       $m^-$ & 0.1 \\
       $\lambda$ & 0.5 \\
       $\theta$ & 0.7 \\
       $d$ & 16 \\
       \bottomrule
       \end{tabular}
   }
\end{table}

\subsection{Setups}

In this section, we describe the necessary Python library and corresponding version for the experiments in the main paper.

\begin{table}[H] 
   \centering
   \resizebox{0.5\linewidth}{!}{
   \begin{tabular}{ll}
       \toprule
       Library & Version  \\ 
       \midrule
       pytorch & 1.12.1  \\
       numpy & 1.24.3 \\
       opencv-python & 4.7.0.72 \\
       pandas & 2.0.2 \\
       pillow & 9.4.0 \\
       torchvision & 0.13.1 \\
       matplotlib & 3.7.1 \\
       icecream & 2.1.3 \\
       seaborn & 0.12.0 \\
       \bottomrule
       \end{tabular}
   }
\end{table}

\section{HouseHolder Orthogonalization}

\subsection{Proof of Lemma 1}
\label{Lemma 1}

\noindent \textbf{Assumption:}

\( V_{l+1} = \{v_{l+1,1}, v_{l+1,2}, \dots, v_{l+1,m}\} \) is a set of \( m \) capsule vectors in layer \( l+1 \) of the network.
$W$ is an orthogonal matrix, i.e. $W^TW=I$.

\noindent \textbf{Proof:}

we aim to prove that the cosine similarity \( t_{ij} \) between any two capsules \( v_{l+1,i} \) and \( v_{l+1,j} \) 
remains constant after multiplied by \( W \).
For an orthogonal matrix $W$, the dot product and vector lengths are preserved.
Let \( \tilde{v}_{l+1,i} = Wv_{l+1,i} \), \( \tilde{v}_{l+1,j} = Wv_{l+1,j} \) denote the transformed vectors. 
Thus, we have:
\begin{equation}
   \begin{aligned}
      \tilde{v}_{l+1,i} \cdot \tilde{v}_{l+1,j} &= (Wv_{l+1,i})^T (W v_{l+1,j}) \\
                                              &=  v_{l+1,i}^T  W^T W  v_{l+1,j} \\
                                              &=  v_{l+1,i}^T  v_{l+1,j}
   \end{aligned}
   \label{equation8}
\end{equation}
and
\begin{equation}
   \begin{aligned}
   \|\tilde{v}_{l+1,i}\|_2 
                              &= \sqrt{( W v_{l+1,i})^T( W v_{l+1,i})} \\
                              &= \sqrt{ v}_{l+1,i}^T  W^T  W  v_{l+1,i} \\
                              &= \sqrt{ v}_{l+1,i}^T  v_{l+1,i} \\
                              &= \| v_{l+1,i}\|_2
   \end{aligned}
   \label{equation9}
\end{equation}
Thus, we obtain transformed cosine similarity $\tilde{t}_{ij}$ in \cref{equation10}:
\begin{equation}
   \begin{aligned}
      \tilde{t}_{ij} &= \frac{(\tilde{v}_{l+1,i}) \cdot (\tilde{v}_{l+1,j})}{\|\tilde{v}_{l+1,i}\|_2 \|\tilde{v}_{l+1,j}\|_2}\\ 
             &= \frac{v_{l+1,i} \cdot v_{l+1,j}}{\| v_{l+1,i}\|_2 \| v_{l+1,j}\|_2}\\
               &= t_{ij}
   \end{aligned}
   \label{equation10}
\end{equation}
Therefore, orthogonal transformations preserve the cosine similarity between any two vectors.

\subsection{Proof of Lemma 3}
\label{HouseHolder Proof} 

\noindent \textbf{Proof:}

Let $W$ represent one of $W_Q$,$W_K$,$W_V$ as $W$ can be expressed as
\begin{equation} \label{householderMulti}
	W = H_0H_1 \dots H_{d-1}
\end{equation}
where $H_i = I- 2a_ia_i^T$. We have
\begin{equation} \label{2orthogonal}
	W^TW = H_{d-1}^T \dots H_1^T H_0^T H_0H_1 \dots H_{d-1}
\end{equation}

We demonstrate that $H_i$ is orthogonal, i.e. $H_i^T H_i = I$. This is obvious, as
\begin{equation} \label{2householder}
	\begin{aligned}
		H_i^T H_i &= (I - 2a_ia_i^T)^T (I - 2a_ia_i^T) \\
		          &= I - 4a_ia_i^T + 4 a_i a_i^T = I
	\end{aligned}
\end{equation}

Therefore, Equation (\ref{2orthogonal}) can be written as $W^TW = \underbrace{I \dots I}_d = I$.

\subsection{Householder as a Regularization Technique}
\label{Householder Regularization} 
\vspace*{-5pt}
\begin{figure}[h]
   \centering
   \includegraphics[width=0.4\textwidth]{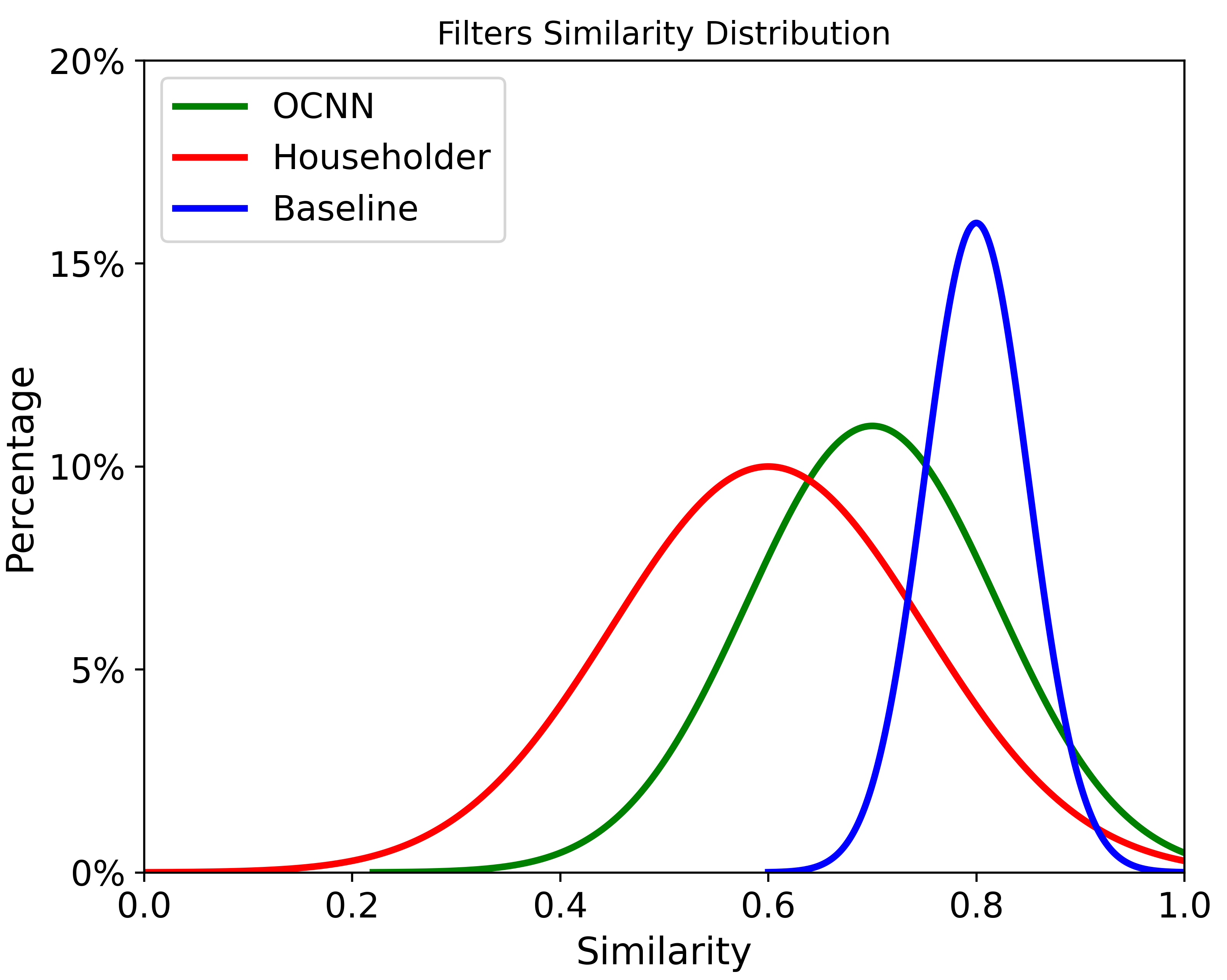}
   \caption{The normalized histogram of pairwise filter similarities in standard ResNet34 with different regularizers. HouseHolder orthogonalization method shows the best performance of descending filter similarity.}   
   \label{orth}
\end{figure}
We demonstrate Householder's role as a regularization technique for neural networks. 
For ResNet18, we flatten and concatenate convolutional kernels into a matrix $W$, 
and orthogonalize it to minimize off-diagonal elements, which reduces channel-wise filter similarity and redundancy. 
To quantify these properties, we used Guided Backpropagation to dynamically visualize the activations\Citep{wang2020orthogonal}. 
Compared to directly computing the covariance matrix of convolutional kernels, 
The gradient-based covariance matrix offers a more comprehensive view of the dynamic behavior of filters. 
We define the gradients from Guided Backpropagation as $G$ and compute its gradient correlation matrix $corr(G)$ as:
\begin{equation}
   \left(\text{diag}(K_{GG})\right)^{-\frac{1}{2}} K_{GG} \left(\text{diag}(K_{GG})\right)^{-\frac{1}{2}}
\end{equation}

where \( K_{GG} = \frac{1}{M} \left( (G - \mathbb{E}[G]) (G - \mathbb{E}[G])^T \right) \), 
$M$ is the number of channels.
\cref{orth} of the off-diagonal elements of 
$corr(G)$ shows a left-shifted distribution for the Householder-regularized model, 
confirming its effectiveness in enhancing filter diversity and reducing redundancy.

\section{Dynamic Routing in Capsule Network}
\label{Dynamic Routing}

\begin{algorithm}
   \caption{Dynamic Routing}
   \label{algo:dynamic_routing}
   \begin{algorithmic}[t]
      \Procedure{ROUTING}{$\hat{u}_{j|i}$, $r$, $l$}
         \For{all capsule $i$ in layer $l$ and capsule $j$ in layer $(l + 1)$} $b_{ij} \leftarrow 0$
         \EndFor
         \For{$T$ iterations}
            \For{all capsule $i$ in layer $l$} $c_i \leftarrow \text{softmax}(b_i)$ 
            \EndFor
            \For{all capsule $j$ in layer $(l + 1)$} $s_j \leftarrow \sum_i c_{ij} \hat{u}_{j|i}$
            \EndFor
            \For{all capsule $j$ in layer $(l + 1)$} $v_j \leftarrow \text{squash}(s_j)$ 
            \EndFor
            \For{all capsule $i$ in layer $l$ and capsule $j$ in layer $(l + 1)$} $b_{ij} \leftarrow b_{ij} + \hat{u}_{j|i} \cdot v_j$
            \EndFor
         \EndFor
         \State \Return $v_j$
      \EndProcedure
   \end{algorithmic}
\end{algorithm}

Algorithm \ref{algo:dynamic_routing} describes the dynamic routing algorithm. 
This algorithm allows lower-level capsule output vectors to be allocated to higher-level capsules based on their similarity, 
thereby achieving an adaptive feature combination. 
However, as evident from $\sum_i c_{ij} \hat{u}_{j|i}$,
each higher-level capsule is a weighted sum of lower-level capsules.
The higher-level capsules are fully connected with the lower level.
Furthermore, the routing algorithm fundamentally serves as an unsupervised clustering process for capsules, 
requiring $r$ iterations to converge the coupling coefficients $c$. 
It's crucial to strike a balance in choosing $r$: an inadequate number of iterations may hinder convergence of $c$, impairing routing efficacy, 
while an excessive count increases computational demands.

In Conclusion, it is crucial to introduce a straightforward, iterative-free routing algorithm.

\end{document}